\documentclass[11pt,a4paper]{article}
\usepackage[hyperref]{naaclhlt2019}
\usepackage{times}
\usepackage{latexsym}
\usepackage{graphicx}
\usepackage{booktabs}
\usepackage{multirow}
\usepackage{amsmath}
\usepackage{url}
\usepackage{subfig}

\aclfinalcopy 

\title{You May Not Need Attention }

\author{
	Ofir Press$^\spadesuit$ \quad
	Noah A. Smith$^{\spadesuit\clubsuit}$ \\
	 $^\spadesuit$Paul G. Allen School of Computer Science \& Engineering,
	University of Washington \\
	$^\clubsuit$Allen Institute for Artificial Intelligence \\
	{\tt \{ofirp,nasmith\}@cs.washington.edu }
}

\date{}

\begin{document}
\maketitle
\begin{abstract}
In NMT, how far can we get without attention and without separate encoding and decoding?
To answer that question, we introduce a recurrent neural translation model that does not use attention and does not have a separate encoder and decoder.  Our \textbf{eager translation model} is low-latency, writing target tokens as soon as it reads the first source token, and uses constant memory during decoding.  It performs on par with the standard attention-based model of \citet{aligntranslate}, and better on long sentences.\footnote{Our code is available at \url{https://github.com/ofirpress/YouMayNotNeedAttention}}
\end{abstract}

\section{Introduction}

Nearly all actively-researched NMT models have the following properties:
\begin{itemize}
\item The decoder uses an attention mechanism over the source sequence representations.~\cite{aligntranslate,luong, transformer}.
\item The encoder and decoder are two different modules, and the encoder must finish encoding the source sentence before the decoder starts operating~\cite{seq2seq,choencdec,aligntranslate,luong,bytenet, transformer}.

\end{itemize}

Here we investigate how well an NMT model can do without these properties.  

To that end, we start with the model of \citet{aligntranslate}, remove the attention mechanism, and unify the encoder and decoder into a single, \emph{simpler} model that resembles the language model of \citet{zaremba}.  The result is that our model can ``eagerly'' begin emitting a translation as soon as it reads the first source word, and can finish translating soon after the last source word is read. 

The \textbf{eager translation model} uses a constant amount of memory, since it needs to use only one previous hidden state (rather than all previous hidden states) at every timestep. Instead of ``cramming a whole sentence into a single vector'', our approach crams a prefix of a source sentence (and its resulting translation) into a dynamic memory vector, emitting target tokens immediately and every time another source word is read. 

In practice, most of the changes required by our eager translation model affect preprocessing (\S\ref{sec:preproc}).  Experimentally,  we show that our model performs on par with the attention-based machine translation model of~\cite{aligntranslate}. We find that our model outperforms the attention-based model on longer sequences (a known challenge for attention-based models) but is less effective on short sequences (\S\ref{sec:exp}). 

We expect that, in the future, this kind of low-latency and low-memory translation may be attractive in some application settings.

\begin{figure}[t]
\centering
\includegraphics[width=\linewidth]{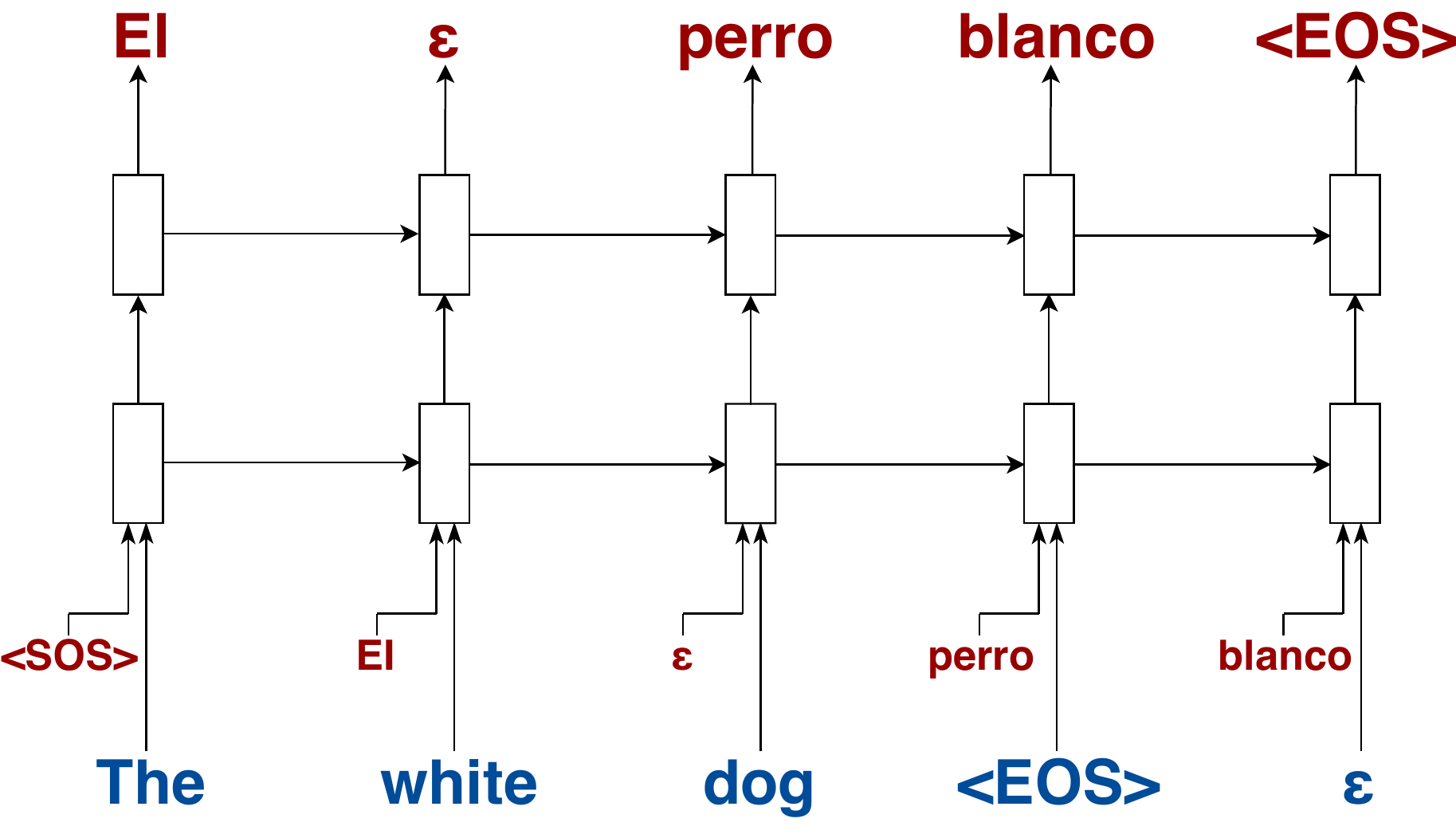}
 \caption{\label{fig:network} The eager model translating the sentence ``The white dog'' into Spanish. Source (target) tokens are in blue (red). $\varepsilon$ is the padding token, which is removed during postprocessing. The diagram presents an eager translation model with two LSTM layers. }
\end{figure}

\section{Data Preprocessing}
\label{sec:preproc}

\begin{figure}[t]
\centering
    \subfloat[]{{\includegraphics[width=0.30\linewidth]{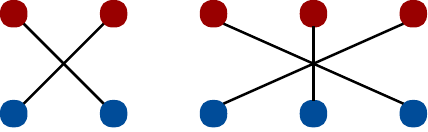} }}%
    \qquad
    \subfloat[]{{\includegraphics[width=0.42\linewidth]{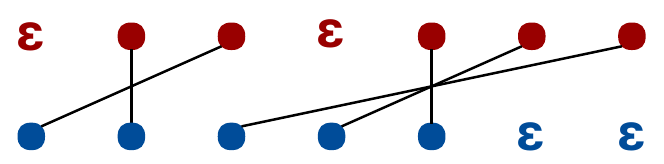} }}%

 \caption{\label{fig:dataproc} A source (blue) and target (red) sequence with their alignment before (a) and after (b) preprocessing to make the pair ``eager feasible.'' }
\end{figure}

\begin{table}[t]
\begin{center}
\begin{tabular}{@{}lcccc@{}}
\toprule
 & {\footnotesize \texttt{FR}$\rightarrow$\texttt{EN}} & {\footnotesize \texttt{EN}$\rightarrow$\texttt{FR}} & {\footnotesize \texttt{DE}$\rightarrow$\texttt{EN}} & {\footnotesize \texttt{EN}$\rightarrow$\texttt{DE}} \\ \midrule
 {\footnotesize  $\varepsilon$ Proportion } & 23\% & 14\% & 23\% & 20\% \\ \bottomrule
\end{tabular}
\end{center}

\caption{\label{epsilon-prop} Average percentage of $\varepsilon$ in the target sentences of the four language directions. Initial padding tokens and padding tokens inserted to make the source and target sequence lengths equal are not counted. }
\end{table}

The eager translation model requires the training data to be preprocessed in a specific way described below.

We begin by inferring a correspondence between words in a source/target sentence pair $(\boldsymbol{s}, \boldsymbol{t})$ assuming each target word is aligned to (at most) one source word, as in~\citet{ibmmodels}.  We describe a source and target sequence as \emph{eager feasible} if, for every aligned pair of words $(s_i, t_j)$, $i \le j$.

Our model requires each source and target sentence in the training set to have this property.  We achieve this by first using the alignments inferred by an off-the-shelf alignment model (\texttt{fast\_align}; \citealp{fastalign}), then inserting the minimal number of $\varepsilon$ (empty) tokens into the target sentence to achieve the desired property. These $\varepsilon$ tokens are used during training and inference, but are removed in a postprocessing step when generating translations.

For example, for the source sentence ``El perro blanco'' (literally: ``The dog white'') the correct translation is ``The white dog''. Assuming the obvious alignment in which the second English word is aligned to the third Spanish word, to make the sequences eager feasible, we modify the target sentence so that it becomes ``The $\varepsilon$ white dog''. A more complex example is given in Fig.~\ref{fig:dataproc}. 

We define the algorithm for making a source and target sequence eager feasible as follows. $\boldsymbol{s} = \langle s_1, \ldots, s_m\rangle$ is the source sentence and $\boldsymbol{t} = \langle t_1, \ldots, t_n\rangle$ is the target sentence.
Let $\mathcal{A}$ be a set of ordered pairs $(i,j)$ such that $t_j$ is aligned to $s_i$. 

We go over the target words one by one, from left to right. Suppose the current target word is $t$, and it currently occupies position $j$ in the target sequence.   If $t$ is aligned to a source word $s_i$ such that $i \le j$, then we move on to the next target word.
Otherwise, we insert enough $\varepsilon$ tokens in the target sequence, just before $t$, so that it shifts to position $i$ in the target sequence.  This, of course, shifts the target words to its right, as well.

Table \ref{epsilon-prop} shows the average percentage of these $\varepsilon$s in the target sequence sets of the four training directions that we trained on.

In order to make the eager model's task  simpler, we also experiment with adding $b \in \{0, 1, \ldots, 5\}$ padding $\varepsilon$ tokens at the beginning of every target sentence. This gives the model a chance to consume more of the source sentence before it begins translating; at inference time, we force the model to produce $b$ $\varepsilon$s before the translation. 
Our preprocessing  algorithm takes these initial padding $\varepsilon$s into account (if they are used), with the result that fewer $\varepsilon$s need to be inserted \emph{between} target tokens.

After the above transformations, if a training pair of sentences do not have the same length, we insert $\varepsilon$s at the end of the shorter sentence to make the lengths equivalent.

\section{Model}
At each timestep, our model first embeds the current input word (in the source language) and the previously selected output word (in the target language) into dense representations both of dimension $E$. These vectors are  concatenated and the resulting vector is then fed into a multi-layered LSTM~\cite{lstm,forgetgate} containing $2E$ units at each layer. 
The output of the LSTM is transformed into a vector of size $E$ using a fully connected layer. The output of that fully connected layer is transformed into a distribution over the target vocabulary using an output embedding matrix and the softmax function. We tie the input embedding matrix of the source language, the input embedding matrix of the target language, and the output embedding matrix~\cite{tying, inantying}.

Our model strongly resembles the recurrent language model of~\citet{zaremba}. As in that model, during training we use teacher forcing~\cite{teacherforcing} and the cross-entropy loss. Note that we treat the padding symbol as a token in the target language, and we do not use a special loss or make any modifications to the objective for timesteps in which the target output is the padding token.

Additionally, unlike most other translation models, our model uses a constant amount of memory during inference. It only has to store the previous hidden state in memory, and does not have to store the representations of all previously encoded words.  
Finally, the decoding complexity of our model is at worse $\mathcal{O}(n+m)$.

\subsection{Aligned Batching}

After preprocessing, all source-target pairs will have equal lengths. We concatenate all the source sentences into a source string and all the target sentences into a target string, keeping them in the same order. This allows us to  train our translation model similarly to how a language model is trained. Specifically, a backpropagation through time (\texttt{BPTT}) hyperparameter is defined. Each element in every batch contains  \texttt{BPTT} source tokens and their respective target tokens. The next batch uses the next  \texttt{BPTT} tokens, and so on. As in language modeling training, the last hidden state from the $(i-1)$th batch becomes the initial hidden state of the $i$th batch.

\subsection{Decoding}
During inference, we use a modified beam search to improve the quality of the outputs. We modify the beam search algorithm as follows:
\begin{itemize}
\item Padding limit: We place an upper limit on the number of padding symbols emitted, by forcing the probability of $\varepsilon$ to zero after the limit is reached. Initial padding symbols are not counted towards this limit. 
\item Source padding injection (SPI): During the development of the model we noticed that the decoder assigns a high probability to the end-of-sequence (EOS) token once the EOS token in the source language is read. We found that translation quality improves if we insert $\varepsilon$ tokens on the \emph{input} side just before the EOS.  If the SPI hyperparameter is set to $c$, our beam search process will consider anywhere from $0$ to $c$ padding $\varepsilon$ tokens before the source EOS.

Source padding injection enables the model to output a sentence that is longer than the input sentence. Without it, the generated sentence length is capped by the source sentence length (and is exactly equal to source sentence length minus the number of generated padding tokens). 
\end{itemize}

\section{Experiments} \label{sec:exp}
\paragraph{Setup} For both EN$\leftrightarrow$FR and EN$\leftrightarrow$DE we train on the WMT 2014\footnote{{\url{http://www.statmt.org/wmt14/translation-task.html}}} dataset, use newstest2013 as the validation dataset, and test on newstest2014. We tokenize the sentences and then segment the words using 32,000 BPE operations~\cite{bpe}. Finally, we shuffle the corpora before training. 

We use four LSTM layers with 1,000 units for our model, and embeddings of size 500. The model is regularized during training using dropout on both the LSTM and the word embeddings, as done by \citet{awd-lstm}. 
We train our model with a batch size of 200, and we backpropagate through time for 60 tokens.  We use SGD and start with a learning rate of 20. We check the perplexity on the validation set every 6,500 updates and halve the learning rate if it does not improve. 
The padding limit, source padding injection value, and beam size that we use for inference on the test set are the ones that perform best on the development set. These values are reported in Table~\ref{PLSPI} in the supplementary material section.

As a reference model we use the OpenNMT~\cite{opennmt} implementation of~\citet{aligntranslate}. We use a model that has two LSTM layers in the encoder and two in the decoder, all with 1,000 units, and embeddings of size 500. This resulted in a model containing a similar number of parameters to our model. The optimization algorithm used is SGD, with a starting rate of 1, which is halved every 10,000 steps if there is no improvement in development set perplexity. 

Both our model and the reference model are trained until there is no improvement on the development set for 50,000 updates. The reference model took 13 hours to train on a single GPU while our models took around 38 hours. Although the eager model can process approximately three times the amount of source tokens per second as the OpentNMT reference model, training takes longer because the eager model requires more epochs to converge. 

We compute BLEU scores on the detokenized outputs using SacreBLEU~\cite{sacrebleu}.

\paragraph{Results}
The results of the eager model are reported in Table~\ref{bleu}.
On FR$\rightarrow$EN and EN$\rightarrow$FR the model is at most 0.8\% lower in terms of BLEU than the reference model. On the harder DE$\rightarrow$EN and EN$\rightarrow$DE tasks, the eager model is at most 4.8\% worse than the reference model. 

Table~\ref{bleu-by-length} breaks down the performance of the best FR$\rightarrow$EN and DE$\rightarrow$EN model by length, showing that the eager model is worse on shorter sequence but better on longer ones, which are known to be difficult for attention-based approaches \cite{sixchallenges}. Table~\ref{blue-by-length-detailed} in the supplementary material section presents a more detailed breakdown of the performance for different source sequence lengths for all four tasks. 

\begin{table}[t]
\begin{center}
\begin{tabular}{@{}lccccc@{}}
\toprule
&   & {\footnotesize \texttt{FR}$\rightarrow$\texttt{EN}} & {\footnotesize \texttt{EN}$\rightarrow$\texttt{FR}} & {\footnotesize \texttt{DE}$\rightarrow$\texttt{EN}} & {\footnotesize \texttt{EN}$\rightarrow$\texttt{DE}} \\ \midrule
\multirow{6}{*}{\begin{tabular}[c]{@{}l@{}}Start\\ $\varepsilon$s\end{tabular}} 
& 0 & 24.42 & 19.97 & 20.11 & 11.50 \\ 
& 1 & 25.76 & 24.81 & 20.81 & 15.81 \\
& 2 & 27.10 & 25.63 & 21.45 & 16.53 \\
& 3 & 27.98 & \textbf{26.98} & 21.39 & 17.36 \\
& 4 & 28.30 & 26.37 & 22.00 & 17.52 \\
& 5 & \textbf{28.47} & 25.49 & \textbf{22.59} & \textbf{17.97} \\ \midrule
\multicolumn{2}{l}{\footnotesize Ref. Model}& 28.56 & 27.20 & 23.01 & 18.89 \\\bottomrule   
\end{tabular}
\end{center}
\caption{\label{bleu} BLEU performance on the test sets for the reference model and our model with zero to five initial $\varepsilon$ padding tokens (as defined in Sec. \ref{sec:preproc})} 
\end{table}

\begin{table}[t]
\begin{center}

\begin{tabular}{@{}lcccc@{}} \toprule
                   & \begin{tabular}[c]{@{}c@{}}Source\\Sentence\\Length\end{tabular} & \begin{tabular}[c]{@{}c@{}}Number\\ of\\ Sentences\end{tabular} & 
                   \begin{tabular}[c]{@{}c@{}}Reference\\Model\\ BLEU\end{tabular} &
                   \begin{tabular}[c]{@{}c@{}}Eager\\Model\\ BLEU\end{tabular}   \\ \midrule
\multirow{5}{*}{\rotatebox[origin=c]{90}{{\footnotesize \texttt{FR}$\rightarrow$\texttt{EN}}}} 
& 1--20 & 864 & \textbf{26.22} & 23.74 \\
 & 21--40 & 1312 & \textbf{29.50} & 29.20 \\
 & 41--60 & 659 & \textbf{28.71} & 27.77 \\
 & 61--80 & 152 & 27.66 & \textbf{27.89} \\
 & 81+ & 16 & 22.10 & \textbf{27.44} \\ \midrule
\multirow{5}{*}{\rotatebox[origin=c]{90}{{\footnotesize \texttt{DE}$\rightarrow$\texttt{EN}}}}  
& 1--20 & 963 & \textbf{22.94} & 20.12 \\
 & 21--40 & 1275 & \textbf{23.07} & 22.95 \\
 & 41--60 & 414 & \textbf{23.06} & 22.53 \\
 & 61--80 & 76 & 23.02 & \textbf{23.51} \\
 & 81+ & 9 & 21.24 & \textbf{24.73} \\ \bottomrule                                   
\end{tabular}
\end{center}

\caption{\label{bleu-by-length} BLEU performance by source sentence length on the FR$\rightarrow$EN and DE$\rightarrow$EN test sets. }
\end{table}

\section{Related Work}

Early models for neural machine translation consisted of a separate encoder and decoder, but without an attention mechanism~\cite{Forcada97, kalch,seq2seq,choencdec}.
Recent state-of-the-art results have been achieved by either recurrent, attention-based neural translation models~\cite{aligntranslate, wu2016google}, transformer-based models~\cite{transformer}, or a combination of these methods~\cite{howmuchattention,bestofbothworlds}.
\citet{bytenet} and \citet{convseq2seq} use convolutional networks in both the encoder and decoder.~\citet{convseq2seq} use an attention mechanism, while~\citet{bytenet} do not.

\citet{raffel2017online} propose a model that uses a monotonic attention mechanism.~\citet{segment2segment} propose a neural transduction model where the alignment is a latent variable. Both of these models have much higher training time requirements than attention-based translation and so only show results on small datasets. 

\citet{npmt} use a reordering layer in their translation model in order to make the input monotonically aligned to its target output. After the reordering step, decoding is done in parallel. The computational complexity of this model is high. 

\citet{neuralHMM} present a recurrent translation model that does not employ an attention mechanism. Their neural HMM model has a decoding complexity of $\mathcal{O}(m^2 n)$ in our notation from \S\ref{sec:preproc}.

All of the aforementioned models use a separate encoder and decoder, and do not begin decoding until the entire source sentence is encoded. In addition, all of the recently introduced models (other than~\citealp{bytenet,npmt,neuralHMM}) use an attention mechanism, either just in the decoder or in both decoder and encoder.   

\citet{elbayad} unify the encoder and decoder but do not use a recurrent architecture, and use an attention mechanism. \citet{gridLSTM} and \citet{bahar2018towards} do not use a typical encoder-decoder architecture, and instead use a 2D LSTM to translate, but these approaches are much slower than encoder-decoder sequence models.

\citet{verbwait} and \citet{gusimul} use reinforcement learning to train online translation models; the latter use an attention-based translation model. In parallel with our work, \citet{ma2018stacl} employ a mechanism similar to our initial padding tokens to improve the performance of their simultaneous translation model. 
While our model does start outputting candidate translations as soon as the first input token is fed in, it is not a simultaneous translation model: we use beam search, which makes it possible (and very probable) for the top candidate translation to change when the next input token is consumed. While these simultaneous translation models have low latency, they do not manage to preform as well as non-simultaneous models.

For morphological inflection generation,~\citet{aharoni} use an alignment model on the training data in order to know when to train the model to output a symbol and when it should read more of the source sequence, similar to our eager feasibility preprocessing.

\section{Conclusion}

We introduce a simple translation model that doesn't use attention and resembles a recurrent language model, and show that it preforms on par with a conventional attention model. 

\bibliography{naaclhlt2019}
\bibliographystyle{acl_natbib}

\appendix

\newpage
\clearpage
\section{Supplementary Material}

\begin{table}[h]
\begin{center}
{\small
\begin{tabular}{@{}lccccc@{}}
\toprule
&   & {\footnotesize \texttt{FR}$\rightarrow$\texttt{EN}} & {\footnotesize \texttt{EN}$\rightarrow$\texttt{FR}} & {\footnotesize \texttt{DE}$\rightarrow$\texttt{EN}} & {\footnotesize \texttt{EN}$\rightarrow$\texttt{DE}} \\ \midrule
\multirow{6}{*}{\begin{tabular}[c]{@{}l@{}}Start\\ $\varepsilon$s\end{tabular}} 
&0 & 5, 4, 25 & 3, 13, 20 & 4, 7, 35 & 5, 7, 25 \\
&1 & 5, 4, 25 & 4, 12, 20 & 4, 7, 35 & 5, 7, 25 \\
&2 & 5, 4, 25 & 3, 14, 20 & 4, 9, 35 & 5, 9, 25 \\
&3 & 6, 5, 35 & 1, 16, 5 & 4, 9, 35 & 4, 9, 25 \\
&4 & 5, 6, 35 & 1, 15, 5 & 3, 9, 35 & 2, 9, 25 \\
&5 & 4, 6, 25 & 2, 15, 5 & 3, 8, 35 & 1, 13, 5 \\ \bottomrule 
\end{tabular}
}
\end{center}
\caption{\label{PLSPI} The padding limit, source padding injection value, and beam size used during inference for each model from Table \ref{bleu}. }
\end{table}

\begin{table}[!htbp]
\begin{center}
{\small
\begin{tabular}{@{}lcccc@{}} \toprule
                   & \begin{tabular}[c]{@{}c@{}}Source\\Sentence\\Length\end{tabular} & \begin{tabular}[c]{@{}c@{}}Number\\ of\\ Sentences\end{tabular} & 
                   \begin{tabular}[c]{@{}c@{}}Reference\\Model\\ BLEU\end{tabular} &
                   \begin{tabular}[c]{@{}c@{}}Eager\\Model\\ BLEU\end{tabular}   \\ \midrule
\multirow{9}{*}{\rotatebox[origin=c]{90}{{\footnotesize \texttt{FR}$\rightarrow$\texttt{EN}}}} 
& 1-10 & 162 & \textbf{23.33} & 12.43 \\
 & 11-20 & 702 & \textbf{26.54} & 25.05 \\
 & 21-30 & 713 & \textbf{29.51} & 28.96 \\
 & 31-40 & 599 & \textbf{29.49} & 29.23 \\
 & 41-50 & 431 & \textbf{28.72} & 27.68 \\
 & 51-60 & 228 & \textbf{28.68} & 27.91 \\
 & 61-70 & 110 & \textbf{28.71} & 28.22 \\
 & 71-80 & 42 & 25.25 & \textbf{27.14} \\
 & 81+ & 16 & 22.10 & \textbf{27.44} \\
 \midrule
\multirow{9}{*}{\rotatebox[origin=c]{90}{{\footnotesize \texttt{EN}$\rightarrow$\texttt{FR}}}} 
& 1-10 & 251 & \textbf{26.31} & 22.32 \\
 & 11-20 & 834 & \textbf{26.15} & 23.74 \\
 & 21-30 & 770 & \textbf{29.33} & 27.88 \\
 & 31-40 & 589 & \textbf{28.37} & 27.49 \\
 & 41-50 & 329 & 26.28 & \textbf{26.85} \\
 & 51-60 & 151 & 26.88 & \textbf{29.63} \\
 & 61-70 & 51 & 19.49 & \textbf{25.46} \\
 & 71-80 & 19 & 22.30 & \textbf{31.64} \\
 & 81+ & 9 & 12.55 & \textbf{26.48} \\
 
\midrule
\multirow{9}{*}{\rotatebox[origin=c]{90}{{\footnotesize \texttt{DE}$\rightarrow$\texttt{EN}}}}  

& 1-10 & 203 & \textbf{20.26} & 12.61 \\
 & 11-20 & 760 & \textbf{23.32} & 21.15 \\
 & 21-30 & 756 & \textbf{22.25} & 22.09 \\
 & 31-40 & 519 & \textbf{23.93} & 23.85 \\
 & 41-50 & 258 & \textbf{23.21} & 22.38 \\
 & 51-60 & 156 & \textbf{22.84} & 22.73 \\
 & 61-70 & 58 & 23.50 & \textbf{23.61} \\
 & 71-80 & 18 & 21.66 & \textbf{23.24} \\
 & 81-+ & 9 & 21.24 & \textbf{24.73} \\
 \midrule
 
 \multirow{9}{*}{\rotatebox[origin=c]{90}{{\footnotesize \texttt{EN}$\rightarrow$\texttt{DE}}}}  
& 1-10 & 211 & \textbf{21.49} & 14.78 \\
 & 11-20 & 859 & \textbf{18.72} & 17.15 \\
 & 21-30 & 781 & \textbf{18.11} & 17.64 \\
 & 31-40 & 492 & \textbf{19.43} & 18.37 \\
 & 41-50 & 227 & \textbf{18.60} & 18.00 \\
 & 51-60 & 116 & \textbf{18.18} & 18.11 \\
 & 61-70 & 38 & 19.87 & \textbf{20.05} \\
 & 71-80 & 6 & \textbf{15.79} & 15.62 \\
 & 81-+ & 7 & \textbf{27.10} & 26.41 \\
 
 \bottomrule                                   
\end{tabular}
}
\end{center}

\caption{\label{blue-by-length-detailed} BLEU performance by source sentence length on all four language directions. }
\end{table}

\end{document}